\documentclass[11pt, a4paper, logo, copyright, nonumbering]{antgroup}
\usepackage[square, numbers]{natbib}
\usepackage{dblfloatfix}
\usepackage{ulem}
\usepackage{caption}
\usepackage{dramatist}
\usepackage{xspace}
\usepackage{pifont}
\usepackage{multirow}

\usepackage{xltabular}
\usepackage{longtable}
\usepackage{hyperref}
\usepackage{amsfonts}
\usepackage{amsmath}
\usepackage{amssymb}
\usepackage{lineno}
\usepackage{multirow}
\usepackage{adjustbox}

\usepackage[bottom]{footmisc}

\usepackage{CJKutf8}
\usepackage{subfigure}
\usepackage{setspace}

\usepackage{dsfont}
\usepackage{array}
\usepackage{tabularx}
\usepackage{subfigure}
\usepackage{xcolor}
\usepackage{tabularx}
\usepackage{booktabs}

\usepackage{lipsum}
\usepackage{multicol}
\usepackage{authblk}
\usepackage{wrapfig}
\usepackage[most,skins,theorems]{tcolorbox}
\usepackage{array}
\usepackage{pifont}
\definecolor{customgray}{RGB}{230,230,230}

\tcbset{
  aibox/.style={
    width=\linewidth,
    top=8pt,
    bottom=4pt,
    colback=gray!10!white,
    colframe=gray!50!black,
    colbacktitle=gray!70!black,
    enhanced,
    center,
    attach boxed title to top left={yshift=-0.1in,xshift=0.15in},
    boxed title style={boxrule=0pt,colframe=white,},
  }
}
\newtcolorbox{AIbox}[2][]{aibox,title=#2,#1}

\makeatletter
\def\@BTrule[#1]{%
  \ifx\longtable\undefined
    \let\@BTswitch\@BTnormal
  \else\ifx\hline\LT@hline
    \nobreak
    \let\@BTswitch\@BLTrule
  \else
     \let\@BTswitch\@BTnormal
  \fi\fi
  \global\@thisrulewidth=#1\relax
  \ifnum\@thisruleclass=\tw@\vskip\@aboverulesep\else
  \ifnum\@lastruleclass=\z@\vskip\@aboverulesep\else
  \ifnum\@lastruleclass=\@ne\vskip\doublerulesep\fi\fi\fi
  \@BTswitch}
\makeatother

\addto\extrasenglish{
}

 {\begin{list}{}%
         {\setlength{\leftmargin}{#1}}%
         \item[]%
 }
 {\end{list}}

\reportnumber{001} %

\renewcommand{\today}{}

\title{\centering DT-Guard: Intent-Driven Reasoning-Active Training for Reasoning-Free LLM Safety Guardrail}

\author{
He Liu$^{*}$,
Changtao Miao$^{*}$,
Xinjie Yang$^{*}$,
Tianle Song,
Yin Wu,
Junchi Chen,
Bintao He,
Xinyuan Zhang,
Bo Zhang$^{\dag}$,
Shi Yan,
Wei Lu,
Wei Wang,
Danyang Xu,
Jiansheng Cai,
Zhe Li
\\
\vspace{-6pt}
Ant Digital Technologies, Ant Group
\vspace{-6pt}
}

\footnotetext[1]{Equal contribution.}
\footnotetext[2]{Corresponding to boyuan.zb@antgroup.com}

\renewcommand{\phi}{\varphi}

\renewcommand{\epsilon}{\varepsilon}
\renewcommand{\imath}{\mathrm{i}}

\newlength{\restsubwidth}
\newlength{\restsubheight}
\newlength{\restsubmoreheight}
\newcommand{\rest}[2]{%
        \settowidth{\restsubwidth}{\ensuremath{#2}}
        \settoheight{\restsubheight}{\ensuremath{{}_{#2}}}
        \ensuremath{{#1\hskip 0.5pt}_{\vrule\kern2pt\parbox[b][%
        4pt][b]{\the\restsubwidth}{%
                        \ensuremath{{}_{#2}}}}}
        }

\begin{abstract}
Large language models deployed in open-world applications require safety guardrails that are both robust to complex risks and efficient enough for low-latency runtime moderation. Existing guardrails face a practical trade-off between lightweight classification-based models, which are efficient but often struggle with concealed intent, ambiguous semantics, and borderline safety decisions, and reasoning-based guards, which improve judgment quality but introduce additional token generation and inference latency.
We present DT-Guard, a content safety guardrail model based on a Reasoning-Active Training, Reasoning-Free Inference paradigm. The key idea is to use reasoning supervision during training while emitting only structured safety labels at inference time. DT-Guard formulates safety judgment as a progressive decision process, Intent → Category → Safety, and constructs an intent-driven dataset with intent labels, risk categories, safety labels, and structured reasoning trajectories. To further improve hard-case robustness, we propose Rollout-Guided Progressive Hard-Case Optimization (RG-PHO), which uses multi-rollout consistency to identify stably mastered, persistently failed, and preference-unstable samples, and applies targeted supervised and preference optimization accordingly.
At inference time, DT-Guard directly generates structured labels without explicit reasoning traces, preserving deployment efficiency. Experiments on prompt-side and response-side safety benchmarks show that DT-Guard achieves average F1 scores of 0.886 and 0.870, respectively. With only a 4B backbone, it reaches a dual-side average F1 of 0.878, outperforming strong 8B guardrail baselines. These results demonstrate that reasoning supervision can be effectively internalized into low-latency safety discrimination.
\end{abstract}

\begin{document}
\maketitle

\begin{figure}[t!]
    \centering
    \includegraphics[width=\linewidth]{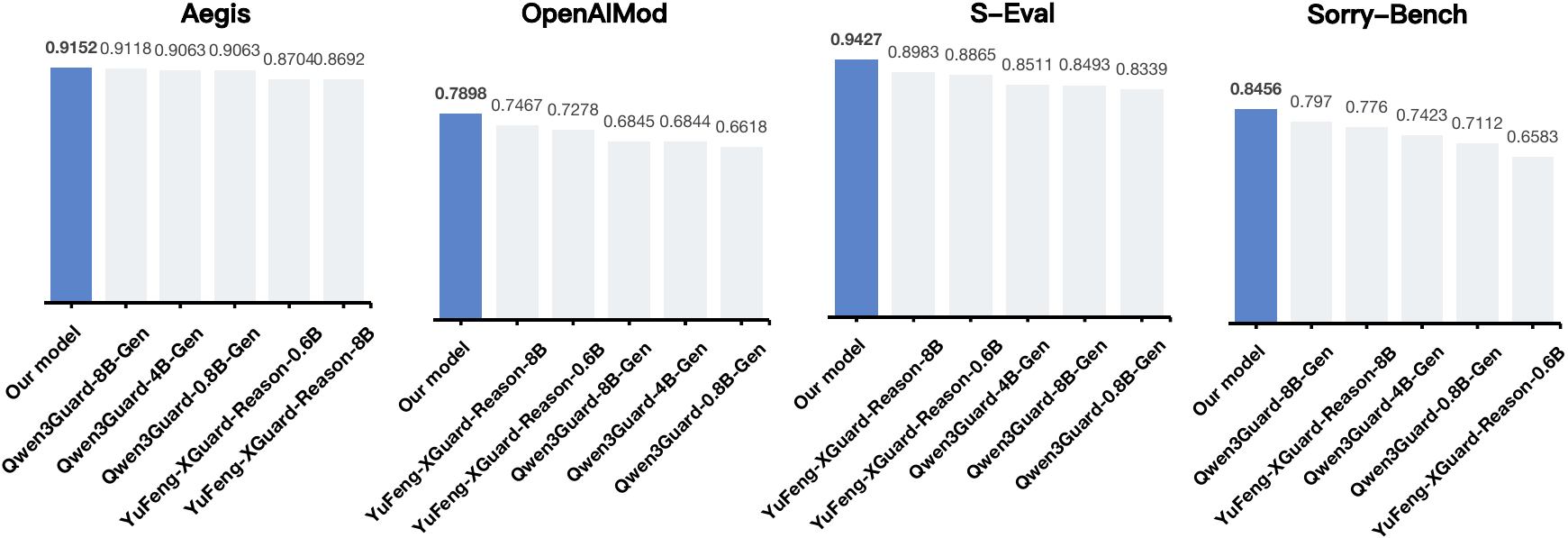}
    \label{fig:data_processing}
    \caption{DT-Guard achieves the top F1 on representative safety benchmarks under reasoning-free inference, outperforming strong guardrail baselines.}
\end{figure}

\section{Introduction}

Large language models (LLMs) have achieved substantial progress in instruction following, knowledge-intensive question answering, complex reasoning, and multi-turn interaction, and are increasingly deployed in open-ended real-world applications~\cite{qwen35blog,singh2026openaigpt5card,li2026lingring26technical}. As their deployment scope expands, LLMs must handle diverse user inputs and generate responses across safety-sensitive scenarios. User inputs may contain concealed harmful intent, adversarial jailbreak attempts, or ambiguous borderline requests, while model outputs may include unsafe, harmful, biased, or policy-violating content under specific contexts~\cite{gcg,autodan}. Safety guardrail models~\cite{llamaguard,shieldgemma,qwen3guard,yufengxguard} have therefore become an important runtime safety layer for detecting and intercepting risks before user inputs are passed to the model or before model responses are returned to users.

Existing guardrail models typically follow one of two inference paradigms. Classification-based guardrails~\cite{llamaguard,shieldgemma,aegis,wildguard,polyguard} directly predict safety labels according to predefined taxonomies and are efficient for real-time deployment. However, they often struggle with concealed intent, ambiguous semantics, and adversarially framed requests, because complex safety judgments are compressed into flat label prediction. Safety risk is not always determined by surface text alone. Requests involving the same sensitive topic may correspond to benign educational inquiry, defensive analysis, risk exploration, or malicious exploitation~\cite{intent-ft,zhuang}. Without explicit intent modeling, guardrails may suffer from over-refusal on benign requests or risk under-detection on malicious ones.

Reasoning-enhanced guardrails provide another direction. By generating explicit chain-of-thought reasoning or explanatory traces, such methods can analyze context more carefully and improve robustness on difficult safety cases~\cite{liu2025guardreasoner,yufengxguard}. Nevertheless, requiring complete reasoning traces or detailed explanations at inference time introduces additional token-generation overhead and response latency, which makes deployment challenging in high-throughput and low-latency industrial systems~\cite{qwen3guard}. These limitations highlight the need for a guardrail training paradigm that can benefit from reasoning supervision while preserving reasoning-free inference.

To address this challenge, we propose \textbf{DT-Guard}, a content safety guardrail model following the \textit{Reasoning-Active Training, Reasoning-Free Inference} paradigm. Instead of treating explicit reasoning as a mandatory deployment-time output, DT-Guard uses reasoning trajectories as supervision signals during training and directly generates structured safety labels during inference. We formulate safety judgment as a progressive decision process,
\textbf{Intent} $\rightarrow$ \textbf{Category} $\rightarrow$ \textbf{Safety}.
The model first identifies the underlying interaction intent, then attributes the relevant risk categories, and finally predicts the safety level. Compared with directly mapping text to a safety label, this structure provides clearer intermediate supervision for difficult cases where surface-level cues alone are unreliable.

Based on this formulation, we construct an intent-driven safety dataset containing prompts, responses, intent labels, risk categories, safety labels, and structured reasoning trajectories. We further introduce \textbf{Rollout-Guided Progressive Hard-Case Optimization} (RG-PHO), which uses multi-rollout consistency to identify different types of residual errors. Persistently failed samples are optimized with stronger supervised correction, while preference-unstable samples are optimized through contrastive preference learning over correct and incorrect rollouts. In this way, explicit reasoning signals are transferred into more stable label-level safety judgment without requiring reasoning-chain generation at inference time.

Experiments on multiple prompt-side and response-side safety benchmarks demonstrate the effectiveness of DT-Guard under Reasoning-Free inference. On 10 prompt-side benchmarks, DT-Guard achieves an average $F_1$ score of $0.886$, improving over Qwen3Guard-8B ($0.852$) by $3.4$ points. On 7 response-side benchmarks, it achieves an average $F_1$ score of $0.870$. Overall, with only a 4B backbone, DT-Guard reaches a dual-side average $F_1$ score of $0.878$, outperforming strong 8B guardrail baselines. These results show that reasoning supervision can be internalized during training and converted into efficient safety discrimination at inference time.

Our main contributions are summarized as follows:
\begin{itemize}
    \item We propose \textbf{DT-Guard}, a safety guardrail model following the \textit{Reasoning-Active Training, Reasoning-Free Inference} paradigm, which internalizes complex safety reasoning without emitting explicit reasoning chains during deployment.
    \item We construct an intent-driven safety judgment framework and dataset, and introduce \textbf{RG-PHO}, a progressive hard-case optimization strategy. It organizes safety discrimination as \textbf{Intent} $\rightarrow$ \textbf{Category} $\rightarrow$ \textbf{Safety}, stratifies hard cases via rollout consistency, and applies Hard-Case SFT and Hard-Case DPO for targeted optimization.
    \item We conduct multiple experiments on input risk detection and response safety moderation, providing an effective training path that balances reasoning capability and deployment efficiency for low-latency safety guardrails.
\end{itemize}

\section{Related Work}

\subsection{Safety Alignment}
The rapid development and deployment of large language models (LLMs) have made safety alignment a central research topic. Existing methods aim to align model behavior with human preferences and safety requirements during training. RLHF~\cite{rlhf} optimizes LLMs with reward models trained from human preference data, while Constitutional AI~\cite{conai} improves harmlessness through self-critique and revision based on predefined principles. DPO~\cite{dpo} further simplifies preference optimization by directly learning from preference pairs. Although these methods improve intrinsic model safety, alignment alone remains insufficient for reliable open-world deployment. Automated jailbreak methods such as GCG~\cite{gcg} and AutoDAN~\cite{autodan} show that aligned models can still be induced to generate unsafe or policy-violating responses. These limitations motivate runtime safety mechanisms that monitor and regulate LLM interactions beyond model-level alignment.

\subsection{Guardrail Models}
Safety guardrails serve as runtime safety layers for moderating user inputs and model outputs. Existing guardrail models usually formulate safety detection as an instruction-following or classification task based on predefined taxonomies. Llama Guard~\cite{llamaguard} trains a dedicated LLM to classify user inputs and model responses according to safety policies. ShieldGemma~\cite{shieldgemma} improves configurability through customizable policy instructions. Aegis~\cite{aegis} introduces fine-grained risk categories, while WildGuard~\cite{wildguard} unifies safety risk detection, jailbreak detection, and refusal detection. PolyGuard~\cite{polyguard} further extends guardrails to multilingual safety moderation. Despite their effectiveness, these methods mainly rely on direct label prediction, providing limited modeling of latent intent and complex decision processes in ambiguous or deceptive cases.

\subsection{Guardrail Reasoning}
Recent studies have explored reasoning-enhanced guardrails to improve robustness and interpretability. GuardReasoner~\cite{liu2025guardreasoner} explicitly models the reasoning process behind safety judgments, while YuFeng-XGuard~\cite{yufengxguard} emphasizes reasoning-centric and interpretable risk perception. However, generating reasoning traces at inference time introduces additional token cost and response latency, making such methods less practical for latency-sensitive applications. To improve deployment efficiency, Qwen3Guard~\cite{qwen3guard} enables token-level real-time monitoring through streaming detection. Nevertheless, existing guardrails still face a tension between reasoning-enhanced judgment and efficient inference. Our DT-Guard addresses this gap by using reasoning and intent supervision during training while preserving reasoning-free structured-label inference during deployment.

\section{Intent-Driven Safety Data Construction}

\begin{figure}[t]
    \centering
    \includegraphics[width=\linewidth]{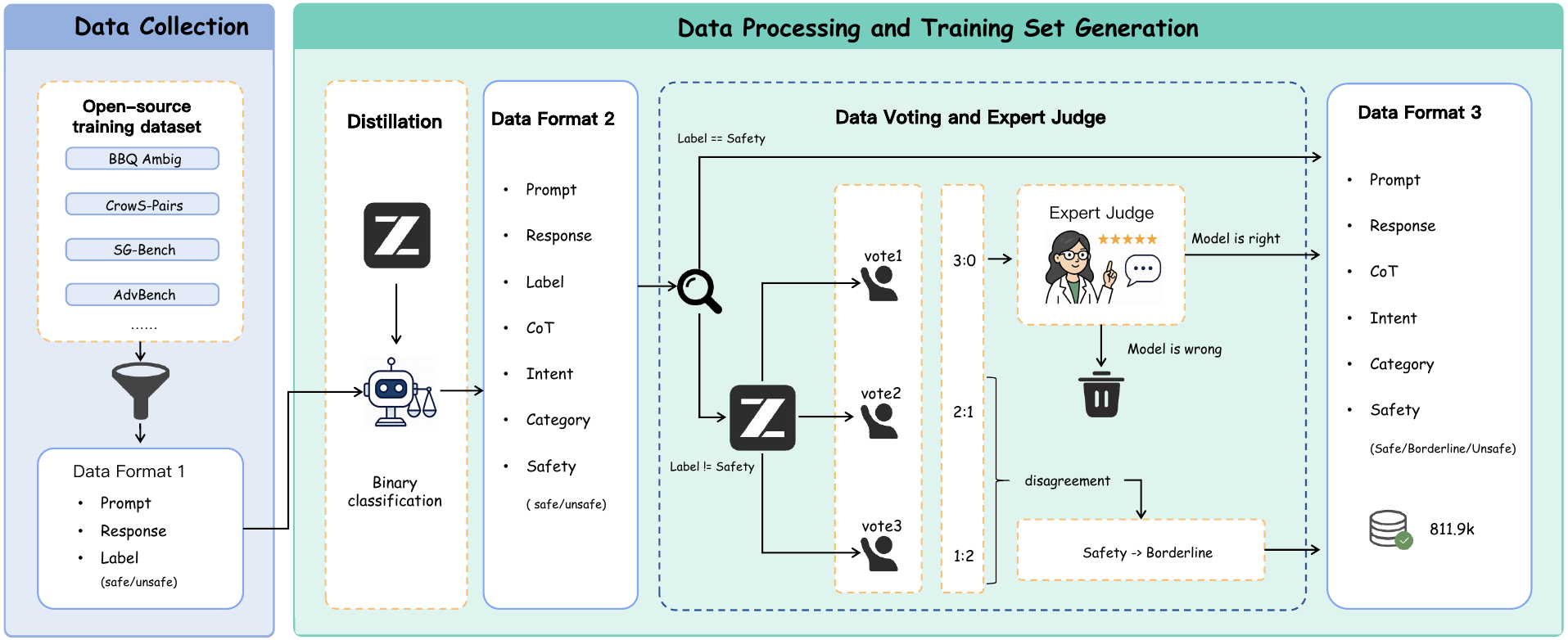}
    \caption{Intent-driven data construction pipeline: heterogeneous safety sources are distilled, filtered by voting, and expert-verified into structured CoT and label supervision.}
    \label{fig:data_construction}
\end{figure}

We build an intent-driven safety corpus for training DT-Guard, as shown in Figure~\ref{fig:data_construction}. Starting from heterogeneous safety resources, we normalize all samples into a unified prompt-response-label format and enrich each instance with chain-of-thought (CoT), intent, risk-category, and safety annotations. This converts safety moderation from flat label prediction into a structured decision path, \textbf{Intent} $\rightarrow$ \textbf{Category} $\rightarrow$ \textbf{Safety}. To reduce annotation noise and expose boundary cases, we combine GLM-5.1 distillation, multi-round voting, and expert verification: agreement cases are retained, confirmed relabels are kept, erroneous cases are filtered, and inconsistent cases are assigned to \textit{Borderline}. The final 811,897-sample corpus provides both structured-label supervision for reasoning-free inference and reasoning trajectories for subsequent RG-PHO training.

\subsection{Data Sources}

We aggregate 1,918,565 raw samples from six safety domains, covering red--blue teaming, jailbreak attacks, alignment data, toxicity, bias, and domain-specific risks. The raw pool contains 949,050 prompt-level samples (49.46\%) for input-side risk detection and 969,515 response-level samples (50.54\%) for output-side safety assessment. After filtering and balancing, the final corpus remains dual-sided, with 450,437 prompt-level samples and 361,460 response-level samples (Table~\ref{tab:raw_data_stats}), enabling comparable supervision for both guardrail entry points.

\begin{table}[t!] \centering \caption{Final corpus scale by task side after filtering and balancing.} \label{tab:raw_data_stats} \begin{tabular}{lrr} \toprule Data Type & Count & Percentage \\ \midrule Prompt-level samples & 450,437 & 55.48\% \\ Response-level samples & 361,460 & 44.52\% \\ \midrule Total samples & 811,897 & 100.00\% \\  \bottomrule \end{tabular} \end{table}

\begin{table}[t]
\centering
\caption{Intent taxonomy and final distribution, separating normal use, risky content, and adversarial attack intent.}
\label{tab:intent_distribution}
\small
\begin{tabular}{lp{8cm}rr}
\toprule
Intent & Definition & Count & Percentage \\
\midrule
Normal & Benign user intent or safe response. & 471,612 & 58.08\% \\
Risky & Unsafe content without attack techniques. & 329,313 & 40.56\% \\
Attack & Unsafe prompt with jailbreak or adversarial attacks. & 10,972 & 1.35\% \\
\midrule
Total & -- & 811,897 & 100.00\% \\
\bottomrule
\end{tabular}
\end{table}

\begin{figure}[t]
    \centering
    \includegraphics[
        width=0.8   \linewidth,
        trim=250pt 210pt 110pt 120pt,
        clip
    ]{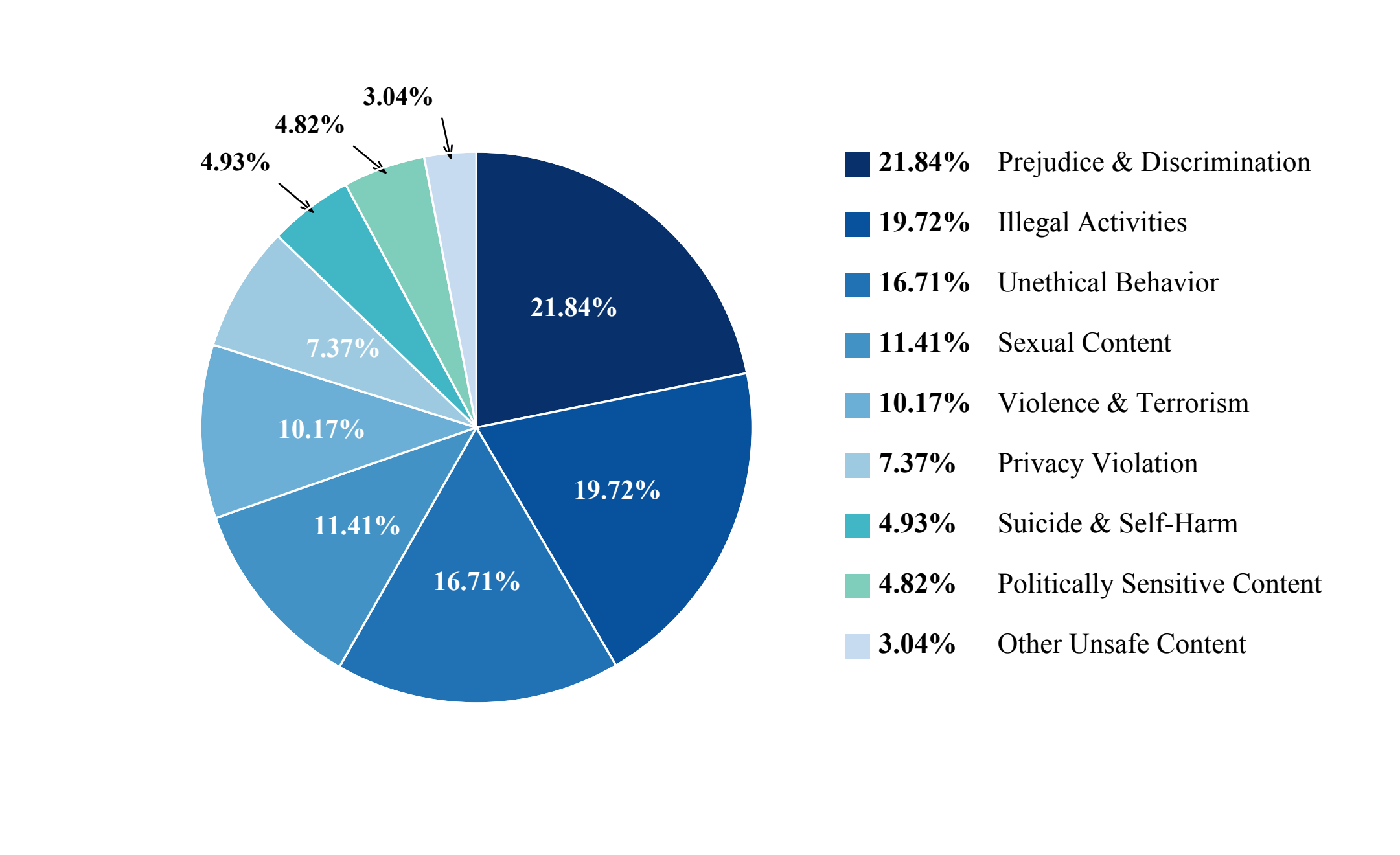}
    \caption{Distribution of unsafe risk categories.}
    \label{fig:unsafe_risk_category_distribution}
\end{figure}

\begin{table}[t]
\centering
\caption{Final safety-label distribution, retaining safe, unsafe, and borderline cases for boundary-risk learning.}
\label{tab:safety_level_distribution}
\small
\begin{tabular}{llrr}
\toprule
Safety Level & Definition & Count & Percentage \\
\midrule
Safe & Benign content & 450,000 & 55.43\% \\
Unsafe & Contains unsafe content & 317,134 & 39.06\% \\
Borderline & Potentially unsafe or ambiguous & 44,763 & 5.51\% \\
\midrule
Total & -- & 811,897 & 100.00\% \\
\bottomrule
\end{tabular}
\end{table}

\subsection{Data Processing and Training Set Construction}

We use a two-stage pipeline to expand annotation dimensions and control label quality.

\textbf{LLM-based Distillation and Quality Filtering.}
Public safety datasets differ in label spaces, risk granularity, and annotation criteria. We therefore use GLM-5.1 as an automatic annotator and require a fixed output schema containing CoT, \textit{Intent}, \textit{Category}, and \textit{Safety}. The intent label follows Table~\ref{tab:intent_distribution}; the category label is selected from the nine risk types in Figure~\ref{fig:unsafe_risk_category_distribution}; and the safety label is one of \textit{Safe}, \textit{Unsafe}, or \textit{Borderline} (Table~\ref{tab:safety_level_distribution}). The annotation order explicitly follows \textit{Intent} $\rightarrow$ \textit{Category} $\rightarrow$ \textit{Safety}, encouraging consistency between interaction motivation, risk attribution, and final adjudication.

\textbf{Multi-round Voting and Expert Verification.}
We next compare the distilled \textit{Safety} label with the original label. Matched samples are retained directly. For mismatched samples, GLM-5.1 performs three additional independent rollouts. Unanimous relabeling results are sent to expert verification; confirmed corrections are kept, and rejected cases are removed. Non-unanimous results (2:1 or 1:2) are treated as annotation-unstable boundary cases and relabeled as \textit{Borderline}. This procedure corrects both false positives that cause over-refusal and false negatives that cause risk under-detection. It retains 1,121,574 high-quality samples from the raw corpus, with a 58.46\% retention rate.

\textbf{Training Set Construction.}
Each retained sample contains dialogue content, CoT reasoning, intent labels, risk categories, and safety labels, providing aligned supervision for intent recognition, risk localization, and safety classification.

\subsection{Data Balancing}

Starting from the 1,121,574 retained samples, we balance the final training distribution along safety labels and risk categories.

\textbf{Safety Label Balancing.}
We target a \textit{Safe}:\textit{Unsafe}:\textit{Borderline} ratio of approximately 5.5:4:0.5. This preserves sufficient unsafe supervision, keeps enough safe samples to reduce over-refusal, and explicitly maintains boundary cases for ambiguity modeling. The final corpus contains 450,000 \textit{Safe} samples (55.43\%), 317,134 \textit{Unsafe} samples (39.06\%), and 44,763 \textit{Borderline} samples (5.51\%), as shown in Table~\ref{tab:safety_level_distribution}.

\textbf{Risk Category Balancing.}
We also perform category-aware sampling over the nine risk types to limit head-category dominance and preserve long-tail coverage. The final distribution covers frequent risks such as \textit{Prejudice and Discrimination} and \textit{Illegal Activities}, while retaining sparse but deployment-critical categories such as \textit{Politically Sensitive Content} and \textit{Others} (Figure~\ref{fig:unsafe_risk_category_distribution}).

After applying both balancing strategies, the final dataset contains 811,897 samples, corresponding to an overall retention rate of 42.32\% relative to the original corpus.

\section{Rollout-Guided Progressive Hard-Case Optimization}

\begin{figure}[t]
    \centering
    \includegraphics[width=\linewidth]{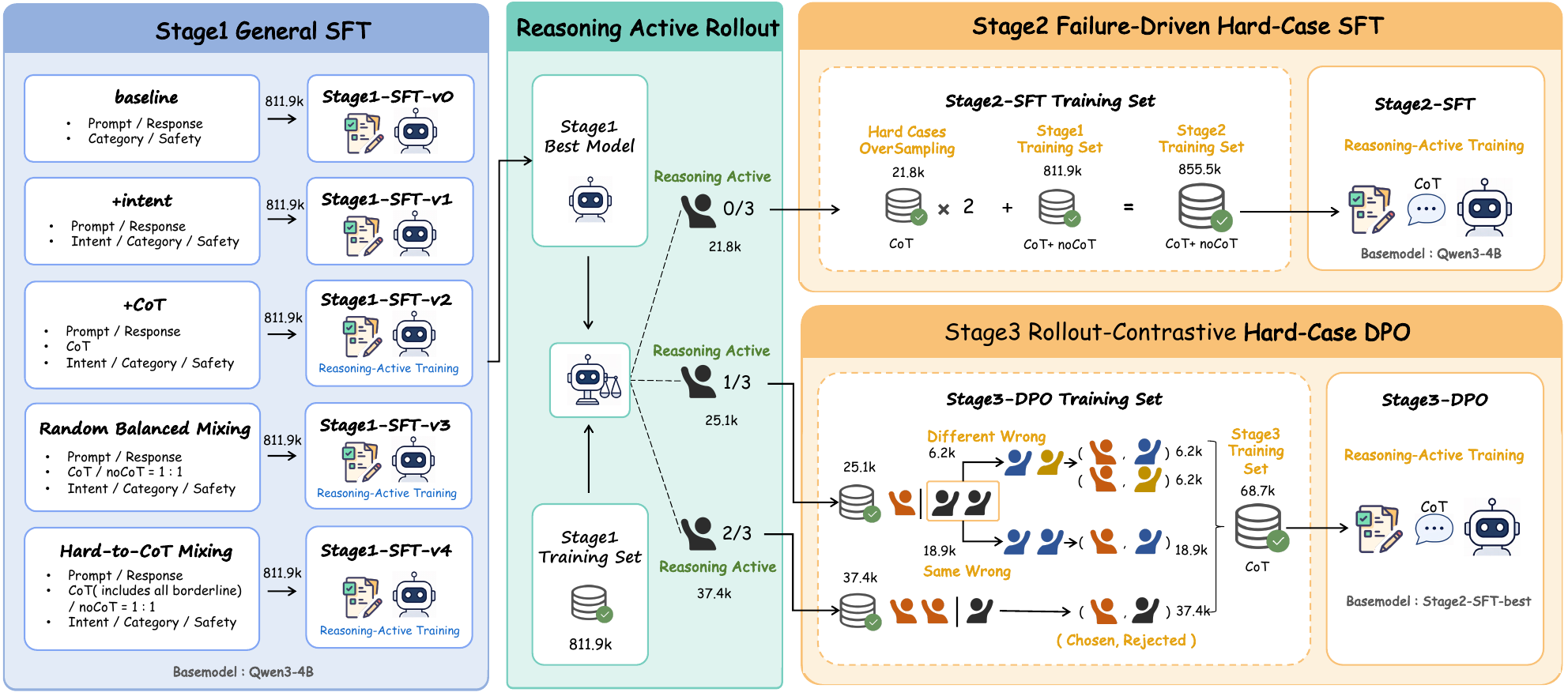}
    \caption{RG-PHO training pipeline: mixed-mode SFT is followed by rollout-based hard-case stratification, hard-case SFT, and rollout-contrastive DPO.}
    \label{fig:training}
\end{figure}

DT-Guard aims to internalize reasoning supervision while keeping inference as structured-label generation. As shown in Figure~\ref{fig:training}, directly fine-tuning on CoT-heavy targets can create a train--test format mismatch: the model learns to depend on explicit reasoning chains, but deployment requires Reasoning-Free outputs. We address this with \textbf{Rollout-Guided Progressive Hard-Case Optimization} (RG-PHO), which uses rollout consistency to estimate sample difficulty and assigns each difficulty pattern to a matching optimization objective.

RG-PHO has three stages. \textit{Intent-Guided Mixed-Mode SFT} first learns the \textbf{Intent} $\rightarrow$ \textbf{Category} $\rightarrow$ \textbf{Safety} decision structure from both structured-label and CoT outputs. The trained model is then rolled out multiple times on training samples to separate stably mastered, persistently failed, and preference-unstable cases. \textit{Failure-Driven Hard-Case SFT} repairs persistently failed samples with stronger supervision, while \textit{Rollout-Contrastive Hard-Case DPO} builds chosen--rejected pairs from correct and incorrect rollouts of preference-unstable samples. This pipeline converts explicit reasoning signals into stable label-level safety judgment.

\subsection{Intent-Guided Mixed-Mode SFT}

The first stage establishes intent-aware safety discrimination. Using the structured taxonomy in Section~3, the model maps each input to \textbf{Intent}, \textbf{Category}, and \textbf{Safety}. \textbf{Intent} captures interaction motivation, \textbf{Category} localize risk types, and \textbf{Safety} gives the final decision.

Unlike standard SFT, this stage mixes two output formats. Borderline and fine-grained cases retain CoT reasoning trajectories to expose intermediate decision logic, while clear safe or unsafe cases more often use compact structured-label outputs. The mixed format injects reasoning supervision without forcing all training samples into an inference-mismatched CoT style.

The training objective is standard autoregressive supervised learning. Given input $x$ and target output $y$, the model minimizes the negative log-likelihood:
\begin{equation}
    \mathcal{L}_{\mathrm{SFT}}
    =
    - \sum_{t=1}^{|y|}
    \log p_{\theta}(y_t \mid x, y_{<t}).
\end{equation}
This stage yields the initial safety model used for rollout-based hard-case identification.

\subsection{Rollout-Consistency-Based Sample Stratification}

After Stage 1, we estimate sample difficulty by prediction stability. For each sample $x_i$, the model generates $K$ independent outputs and produces safety-label predictions. The rollout consistency score is:
\begin{equation}
    s_i
    =
    \sum_{k=1}^{K}
    \mathbf{1}\left[\hat{y}_{i}^{(k)} = y_i\right].
\end{equation}
We use three rollouts in all experiments. Samples with $3/3$ correct predictions are \textbf{stably mastered}; samples with $0/3$ correct predictions are \textbf{persistently failed}; and samples with $1/3$ or $2/3$ correct predictions are \textbf{preference-unstable}. These groups correspond to replay suppression, capability repair, and preference calibration, respectively.

This behavior-driven stratification prevents all residual errors from being optimized identically, and determines whether a sample should be skipped, supervised, or converted into preference pairs.

\subsection{Failure-Driven Hard-Case SFT}

The second stage applies \textit{Failure-Driven Hard-Case SFT} to persistently failed samples. Since the model produces no correct rollout for these cases, DPO lacks a naturally generated chosen response and is poorly matched to the error type.

We therefore use CoT-supervised correction to teach the missing path from text understanding to intent recognition, risk attribution, and safety adjudication. Unlike Stage 1, this stage is not designed to expand coverage; it targets systematic failures that cause risk under-detection or over-refusal.

After Failure-Driven Hard-Case SFT, the model receives new supervision on previously failed samples, providing a more stable initialization for subsequent preference optimization.

\subsection{Rollout-Contrastive Hard-Case DPO}

The third stage applies \textit{Rollout-Contrastive Hard-Case DPO} to preference-unstable samples. These samples already contain both correct and incorrect model-generated outputs, making them natural candidates for preference learning.

For each sample, rollouts consistent with the ground-truth label are chosen responses, and inconsistent rollouts are rejected responses. When multiple distinct incorrect outputs exist, we construct one-to-many preference pairs. Since the outputs retain CoT trajectories, DPO optimizes both the reasoning path and the final safety decision.

The DPO objective is:
\begin{equation}
    \mathcal{L}_{\mathrm{DPO}}
    =
    -
    \mathbb{E}_{(x,y^+,y^-)}
    \log \sigma
    \left(
    \beta
    \left[
    \log \frac{p_{\theta}(y^+ \mid x)}{p_{\mathrm{ref}}(y^+ \mid x)}
    -
    \log \frac{p_{\theta}(y^- \mid x)}{p_{\mathrm{ref}}(y^- \mid x)}
    \right]
    \right),
\end{equation}
where $y^+$ denotes the chosen output, $y^-$ denotes the rejected output, $p_{\mathrm{ref}}$ is the reference model, and $\beta$ controls preference strength. The objective increases the relative probability of correct reasoning paths and safety labels over incorrect adjudications.

Compared with further SFT, DPO is better suited to samples where the model can already generate a correct answer but selects it inconsistently. The contrastive rollout pairs calibrate this selection behavior.

\subsection{Reasoning-Free Inference}

After training, DT-Guard defaults to Reasoning-Free inference. It does not output CoT traces, and directly generates \textbf{Intent}, \textbf{Category}, and \textbf{Safety}. This matches the structured-label format used in mixed-mode training.

Reasoning-Free inference removes explicit reasoning generation, not the reasoning supervision learned during training. RG-PHO converts CoT samples, hard-case SFT, and rollout-contrastive DPO into label-level discrimination capability without adding reasoning-chain generation overhead.

\begin{table*}[t!]
\centering
\caption{Training variants for isolating intent labels, CoT allocation, hard-case SFT, and rollout-contrastive DPO.}
\label{tab:training_variants}
\small
\begin{tabular}{lp{7.8cm}p{4.2cm}}
\toprule
Variant & Model Configuration & Purpose \\
\midrule
Stage1-SFT-v0 &
Qwen3-4B-base; without Intent labels or CoT training. &
Baseline. \\

Stage1-SFT-v1 &
Qwen3-4B-base; with Intent labels, without CoT training. &
Evaluate the effect of Intent supervision. \\

Stage1-SFT-v2 &
Qwen3-4B-base; with Intent labels and CoT training; CoT disabled during inference. &
Evaluate full CoT training. \\

Stage1-SFT-v3 &
Qwen3-4B-base; with Intent labels; mixed noCoT/CoT training data (1:1); CoT disabled during inference. &
Evaluate mixed CoT/noCoT training. \\

Stage1-SFT-v4 &
Qwen3-4B-base; with Intent labels; all Borderline samples trained with CoT, remaining samples mixed at a 1:1 noCoT/CoT ratio; CoT disabled during inference. &
Evaluate selective CoT for Borderline samples. \\

Stage2-SFT &
Stage1-SFT-v4 + SFT with upsampled samples misclassified in all three rollouts. &
Evaluate hard-sample learning. \\

Stage3-DPO &
Stage2-SFT + DPO using synthesized preference pairs from partially misclassified rollout samples. &
Evaluate DPO-based preference optimization. \\
\bottomrule
\end{tabular}
\end{table*}

\section{Experiments and Analysis}

This section evaluates DT-Guard on prompt-side risk detection and response-side safety auditing. We first define the evaluation protocol, then compare against strong guardrail baselines, and finally analyze the contribution of each training component under Reasoning-Free inference.

\subsection{Experimental Setup}

We use Qwen3-4B as the backbone and compare with the Qwen3Guard and YuFeng-XGuard model families. Prompt-side evaluation covers 10 benchmarks, including ToxicChat, OpenAIModeration, AegisSafety, AegisSafety2.0, SimpleSafetyTests, HarmBench-Prompt, WildGuard-Prompt, SafetyEval, Sorry-Bench, and XSTest. Response-side evaluation covers 7 benchmarks, including HarmBench-Response, SafeRLHF, BeaverTails, XSTest-Response, AegisSafety2.0-Response, WildGuard-Response, and Think.

All experiments report classification $F_1$. We summarize prompt-side average, response-side average, and dual-side average $F_1$. For baselines with multiple decision settings, we report the best setting on each benchmark.
\begin{table*}[t]
\centering
\small
\caption{Prompt-side $F_1$ comparison on 10 safety benchmarks. The best result in each column is highlighted in bold.}
\label{tab:main_results}
\resizebox{\textwidth}{!}{
\begin{tabular}{llccccccccccc}
\toprule
\textbf{Model} &
\textbf{Params} &
\textbf{Toxic} &
\textbf{OpenAI} &
\textbf{Aegis} &
\textbf{Aegis2} &
\textbf{SimpST} &
\textbf{HarmB} &
\textbf{WildG} &
\textbf{S-Eval} &
\textbf{Sorry} &
\textbf{XSTest} &
\textbf{Avg.} \\
\midrule
\multicolumn{13}{c}{\textit{Existing Guard Models}} \\
\midrule
Qwen3Guard-0.8B-Gen & 0.8B & 0.578 & 0.662 & 0.906 & 0.851 & 0.990 & 0.988 & 0.878 & 0.834 & 0.711 & 0.853 & 0.825 \\
Qwen3Guard-4B-Gen   & 4B   & 0.616 & 0.684 & 0.906 & 0.858 & 0.995 & \textbf{1.000} & 0.887 & 0.851 & 0.742 & 0.899 & 0.844 \\
Qwen3Guard-8B-Gen   & 8B   & 0.617 & 0.685 & 0.912 & 0.861 & 0.995 & \textbf{1.000} & \textbf{0.891} & 0.849 & 0.797 & 0.908 & 0.852 \\
YuFeng-XGuard-Reason-0.6B & 0.6B & 0.713 & 0.728 & 0.870 & \textbf{0.862} & 0.995 & 0.824 & 0.879 & 0.887 & 0.658 & 0.918 & 0.833 \\
YuFeng-XGuard-Reason-8B & 8B & 0.696 & 0.747 & 0.869 & 0.854 & \textbf{1.000} & 0.821 & 0.877 & 0.898 & 0.776 & \textbf{0.954} & 0.849 \\
\midrule
\multicolumn{13}{c}{\textit{DT-Guard (Ours)}} \\
\midrule
Stage1-SFT-v0 & 4B & 0.658 & 0.740 & 0.913 & 0.830 & \textbf{1.000} & 0.947 & 0.870 & 0.907 & 0.816 & 0.871 & 0.855 \\
Stage1-SFT-v1 & 4B & 0.665 & 0.768 & 0.910 & 0.844 & \textbf{1.000} & 0.995 & 0.870 & 0.909 & 0.798 & 0.874 & 0.863 \\
Stage1-SFT-v2 & 4B & 0.656 & \textbf{0.813} & 0.875 & 0.836 & \textbf{1.000} & 0.805 & 0.829 & 0.889 & 0.757 & 0.847 & 0.831 \\
Stage1-SFT-v3 & 4B & 0.638 & 0.758 & 0.902 & 0.837 & \textbf{1.000} & 0.866 & 0.864 & 0.902 & 0.772 & 0.860 & 0.840 \\
Stage1-SFT-v4 & 4B & \textbf{0.723} & 0.803 & 0.850 & 0.851 & \textbf{1.000} & 0.998 & 0.888 & 0.842 & 0.702 & 0.936 & 0.859 \\
Stage2-SFT & 4B & 0.660 & 0.798 & 0.876 & 0.843 & \textbf{1.000} & \textbf{1.000} & 0.884 & 0.901 & 0.818 & 0.867 & 0.865 \\
\textbf{Stage3-DPO} & 4B & 0.697 & 0.790 & \textbf{0.915} & 0.847 & \textbf{1.000} & \textbf{1.000} & 0.888 & \textbf{0.943} & \textbf{0.846} & 0.929 & \textbf{0.886} \\
\bottomrule
\end{tabular}}
\end{table*}

\begin{table*}[t]
\centering
\small
\caption{Response-side $F_1$ comparison on 7 safety benchmarks. The best result in each column is highlighted in bold.}
\label{tab:response_results}
\resizebox{\textwidth}{!}{
\begin{tabular}{llcccccccc}
\toprule
\textbf{Model} &
\textbf{Params} &
\textbf{HarmB} &
\textbf{SafeRLHF} &
\textbf{BeaverTails} &
\textbf{XSTest} &
\textbf{Aegis2} &
\textbf{WildG} &
\textbf{Think} &
\textbf{Avg.} \\
\midrule
\multicolumn{10}{c}{\textit{Existing Guard Models}} \\
\midrule
Qwen3Guard-0.8B-Gen & 0.8B & \textbf{0.985} & 0.806 & 0.862 & 0.892 & 0.844 & 0.761 & 0.859 & 0.858 \\
Qwen3Guard-4B-Gen   & 4B   & 0.964 & 0.812 & 0.860 & 0.881 & 0.849 & 0.758 & 0.845 & 0.853 \\
Qwen3Guard-8B-Gen   & 8B   & 0.947 & 0.820 & \textbf{0.865} & \textbf{0.921} & \textbf{0.862} & 0.782 & 0.842 & 0.863 \\
YuFeng-XGuard-Reason-0.6B & 0.6B & 0.802 & 0.848 & 0.849 & 0.872 & 0.819 & 0.780 & 0.872 & 0.835 \\
YuFeng-XGuard-Reason-8B & 8B & 0.817 & \textbf{0.857} & 0.849 & 0.885 & 0.832 & 0.779 & 0.871 & 0.841 \\
\midrule
\multicolumn{10}{c}{\textit{DT-Guard (Ours)}} \\
\midrule
Stage1-SFT-v0 & 4B & 0.931 & 0.819 & 0.853 & 0.897 & 0.825 & 0.750 & 0.853 & 0.847 \\
Stage1-SFT-v1 & 4B & 0.936 & 0.818 & 0.856 & 0.874 & 0.832 & 0.769 & 0.854 & 0.848 \\
Stage1-SFT-v2 & 4B & 0.925 & 0.771 & 0.832 & 0.847 & 0.831 & 0.746 & 0.818 & 0.824 \\
Stage1-SFT-v3 & 4B & 0.925 & 0.830 & 0.850 & 0.838 & 0.815 & 0.749 & 0.865 & 0.839 \\
Stage1-SFT-v4 & 4B & 0.953 & 0.826 & 0.854 & 0.898 & 0.834 & 0.784 & 0.871 & 0.860 \\
Stage2-SFT & 4B & 0.964 & 0.825 & 0.855 & 0.909 & 0.833 & 0.783 & 0.871 & 0.863 \\
\textbf{Stage3-DPO} & 4B & 0.964 & 0.826 & 0.856 & 0.917 & 0.847 & \textbf{0.798} & \textbf{0.884} & \textbf{0.870} \\
\bottomrule
\end{tabular}}
\end{table*}

\begin{table}[t]
\centering
\small
\caption{Overall average $F_1$ comparison across prompt-side and response-side benchmarks.}
\label{tab:overall}
\begin{tabular}{llccc}
\toprule
\textbf{Model} &
\textbf{Params} &
\textbf{Prompt} &
\textbf{Response} &
\textbf{Avg.} \\
\midrule
\multicolumn{5}{c}{\textit{Existing Guard Models}}\\
\midrule
Qwen3Guard-0.8B-Gen & 0.8B & 0.825 & 0.858 & 0.842 \\
Qwen3Guard-4B-Gen   & 4B   & 0.844 & 0.853 & 0.848 \\
Qwen3Guard-8B-Gen   & 8B   & 0.852 & 0.863 & 0.858 \\
YuFeng-XGuard-Reason-0.6B & 0.6B & 0.833 & 0.835 & 0.834 \\
YuFeng-XGuard-Reason-8B & 8B & 0.849 & 0.841 & 0.845 \\
\midrule
\multicolumn{5}{c}{\textit{DT-Guard (Ours)}}\\
\midrule
Stage1-SFT-v0 & 4B & 0.855 & 0.847 & 0.851 \\
Stage1-SFT-v1 & 4B & 0.863 & 0.848 & 0.856 \\
Stage1-SFT-v2 & 4B & 0.831 & 0.824 & 0.828 \\
Stage1-SFT-v3 & 4B & 0.840 & 0.839 & 0.840 \\
Stage1-SFT-v4 & 4B & 0.859 & 0.860 & 0.860 \\
Stage2-SFT & 4B & 0.865 & 0.863 & 0.864 \\
\textbf{Stage3-DPO} & \textbf{4B} & \textbf{0.886} & \textbf{0.870} & \textbf{0.878} \\
\bottomrule
\end{tabular}
\end{table}

\subsection{Main Results}

Table~\ref{tab:main_results} and Table~\ref{tab:response_results} show that DT-Guard improves both input-side and output-side safety classification. On the prompt side, Stage3-DPO reaches an average $F_1$ of $0.886$, outperforming Qwen3Guard-8B-Gen ($0.852$) and YuFeng-XGuard-Reason-8B ($0.849$). On the response side, it reaches $0.870$, exceeding Qwen3Guard-8B-Gen($0.863$) and YuFeng-XGuard-Reason-8B ($0.841$).

Across both task types, DT-Guard obtains a dual-side average $F_1$ of $0.878$ (Table~\ref{tab:overall}), surpassing Qwen3Guard-8B-Gen ($0.858$) and YuFeng-XGuard-Reason-8B ($0.845$) with only a 4B backbone. The gain therefore comes primarily from intent-driven supervision and RG-PHO rather than model scale.

\begin{table}[t]
\centering
\caption{CoT allocation ablation under Reasoning-Free inference. Selective CoT on Borderline samples gives the best average $F_1$.}
\label{tab:cot_ablation}
\small
\begin{tabular}{lccc}
\toprule
 \textbf{Training Strategy} &
\textbf{Prompt} &
\textbf{Response} &
\textbf{Avg.} \\
\midrule
NoCoT only & 0.863 & 0.848 & 0.856 \\
Full CoT & 0.831 & 0.824 & 0.828 \\
Mixed CoT/NoCoT (1:1) & 0.840 & 0.839 & 0.840 \\
Borderline CoT + Mixed (1:1) & \textbf{0.859} & \textbf{0.860} & \textbf{0.860} \\
\bottomrule
\end{tabular}
\end{table}

\begin{table}[t]
\centering
\small
\caption{Progressive average $F_1$ gains of DT-Guard across the three-stage training pipeline. Gain is measured over Stage1-SFT-v0.}
\label{tab:training_progress}
\begin{tabular}{llcccc}
\toprule
\textbf{Stage} & \textbf{Variant} & \textbf{Prompt} & \textbf{Response} & \textbf{Avg.} & \textbf{Gain} \\
\midrule
Baseline & v0 & 0.855 & 0.847 & 0.851 & -- \\
\midrule
\multirow{4}{*}{Stage 1}
& +Intent Labels & 0.863 & 0.848 & 0.856 & +0.5 \\
& +Full CoT & 0.831 & 0.824 & 0.828 & $-$2.3 \\
& +Mixed CoT/NoCoT (1:1) & 0.840 & 0.839 & 0.840 & $-$1.1 \\
& +Borderline CoT & 0.859 & 0.860 & 0.860 & +0.9 \\
\midrule
Stage 2 & +Hard-example SFT & 0.865 & 0.863 & 0.864 & +1.3 \\
Stage 3 & +Preference-aware DPO & \textbf{0.886} & \textbf{0.870} & \textbf{0.878} & \textbf{+2.7} \\
\bottomrule
\end{tabular}
\end{table}

\subsection{Ablation Study and Training Analysis}

We ablate intent labels, CoT allocation, hard-case SFT, and DPO to isolate the contribution of each RG-PHO component. The results show that gains come from matching supervision type to sample difficulty, rather than from adding CoT uniformly.

\subsubsection{Effect of Intent Labels}

Intent supervision models interaction motivation before the final safety decision. Compared with Stage1-SFT-v0, adding \textbf{Intent} improves prompt-side average $F_1$ from $0.855$ to $0.863$, with a smaller but positive response-side gain.

This gain is largest on prompt-side tasks, where semantically similar requests may reflect normal use, risky exploration, or attack intent. Intent labels provide an intermediate constraint before category attribution and safety prediction.

\subsubsection{Effect of CoT Training Strategy}

Table~\ref{tab:cot_ablation} shows that CoT supervision is useful only when allocated selectively. Full CoT training reduces dual-side average $F_1$ from $0.856$ to $0.828$, indicating a format mismatch between CoT-heavy training and Reasoning-Free inference.

Random CoT/noCoT mixing partially alleviates the mismatch but remains below the noCoT baseline. In contrast, assigning CoT to Borderline samples and mixing the remaining data improves the average to $0.860$, supporting our selective reasoning-supervision design.

\subsubsection{Rollout-Consistency-Based Sample Stratification}

Rollout consistency separates residual errors by model behavior. Stably mastered samples offer limited marginal value for replay; persistently failed samples lack a correct generated path and require supervised repair; preference-unstable samples already contain correct and incorrect rollouts and are suitable for DPO. This stratification is the routing mechanism that decides whether a sample contributes to Stage2-SFT or Stage3-DPO.

\subsubsection{Effect of Hard-Case SFT}

Failure-Driven Hard-Case SFT targets persistently failed samples instead of replaying the full dataset uniformly. As shown in Table~\ref{tab:training_progress}, it improves the dual-side average $F_1$ from $0.860$ to $0.864$.

Although the gain is smaller than DPO, this stage creates a stronger initialization for preference optimization. For samples without any correct rollout, supervised correction is more appropriate than preference ranking because no model-generated chosen response exists.

\subsubsection{Effect of Hard-Case DPO}

Rollout-Contrastive Hard-Case DPO provides the largest single-stage gain. It constructs chosen--rejected pairs from correct and incorrect rollouts of preference-unstable samples, encouraging the model to consistently select the correct adjudication path.

As shown in Table~\ref{tab:training_progress}, DPO improves the dual-side average $F_1$ from $0.864$ to $0.878$. The prompt-side average increases from $0.865$ to $0.886$, indicating that preference optimization is especially effective for boundary requests, jailbreak-style inputs, and over-refusal control.

\subsection{Analysis of Reasoning-Free Inference}

All main results are obtained under Reasoning-Free inference, where DT-Guard outputs only structured safety labels. Unlike explicit reasoning-based guardrails, our model uses reasoning primarily as training supervision through mixed SFT, hard-case SFT, and rollout-contrastive DPO.

The ablations support this asymmetry. Full CoT training degrades when CoT inference is disabled, whereas selective CoT and RG-PHO improve performance. Reasoning enhancement therefore depends less on emitting reasoning chains at deployment, and more on aligning reasoning supervision with the Reasoning-Free output format.

\section{Conclusion}

We present \textbf{DT-Guard}, a content safety guardrail model designed for low-latency deployment. Unlike guardrail methods that rely on explicit chain-of-thought or explanation generation during inference, DT-Guard follows a \textit{Reasoning-Active Training, Reasoning-Free Inference} paradigm: explicit reasoning supervision is used during training, while inference outputs only structured safety labels. This enables the model to internalize intent recognition, risk attribution, and safety adjudication required for complex safety judgment while preserving an efficient inference format.
To achieve this, we construct an intent-driven safety data framework that organizes safety discrimination as \textbf{Intent} $\rightarrow$ \textbf{Category} $\rightarrow$ \textbf{Safety}, and introduce \textbf{Rollout-Guided Progressive Hard-Case Optimization}. This training strategy uses multi-rollout consistency to identify different types of hard cases, and improves the model's judgment on persistently failed and preference-unstable samples through Hard-Case SFT and Hard-Case DPO, respectively. Experiments show that DT-Guard achieves consistent gains on both prompt-side and response-side safety benchmarks, outperforming several 8B guardrail models with only 4B parameters. These results validate that reasoning supervision during training can be effectively converted into low-latency discrimination capability at inference time.

\clearpage
\bibliographystyle{unsrtnat} 
\bibliography{main}

\end{document}